\documentclass[journal]{IEEEtran}
\IEEEoverridecommandlockouts
\usepackage{cite}
\usepackage{amsmath,amssymb,amsfonts}
\usepackage[ruled,vlined]{algorithm2e}
\usepackage{graphicx}
\usepackage{textcomp}
\usepackage{xcolor}
\def\BibTeX{{\rm B\kern-.05em{\sc i\kern-.025em b}\kern-.08em
    T\kern-.1667em\lower.7ex\hbox{E}\kern-.125emX}}

\usepackage{float}

\begin{document}

\title{A Parallel Approach for Real-Time Face Recognition from a Large Database}

\author{\IEEEauthorblockN{Ashish Ranjan*,
Varun Nagesh Jolly Behera and
Motahar Reza}

\IEEEauthorblockA{R\&D Division,\\
Just Another Media Laboratory (JAM Lab)\\
Mumbai, India\\
Email: ashish.ranjan@jamlab.in, varun.behera@jamlab.in, mreza@gitam.edu}}

\maketitle

\begin{abstract}
We present a new facial recognition system, capable of identifying a person, provided their likeness has been previously stored in the system, in real time. The system is based on storing and comparing facial embeddings of the subject, and identifying them later within a live video feed. This system is highly accurate, and is able to tag people with their ID in real time. It is able to do so, even when using a database containing thousands of facial embeddings, by using a parallelized searching technique. This makes the system quite fast and allows it to be highly scalable.
\end{abstract}


\section{Introduction}
Current implementations of video surveillance systems are designed for purposes such as movement detection and future enquiries.\cite{hampapur2003smart} These systems require human insight to actually use the huge amounts of data generated by video surveillance. Due to the exponential size of this data, and the benefits of using this data to assist an early warning system, or may be used as part of some other inference engine. Also it is impossible for a human being to skim through hours of video data without missing key details, which a computer vision system may easily identify\cite{robertson2008automatic}. Such a system can accomplish the same tasks that may be given to a human being, faster and more gracefully. One major part of a surveillance system is its facial recognition engine. The reason for using such a system would be primarily to identify individuals that may have been involved in some sort of unlawful activity. Other uses may be biometric verification\cite{maas2014offering}, tracking a specified individuals movements, terrorism prevention systems\cite{srivastava2017safety}, etc.

\section{Goals}
We aim to develop a facial recognition system, able to process and identify individuals in real time. The system will be capable of identifying people from existing information and have the ability to add new faces to its database. The entire search process will be done in parallel, allowing for faster information processing and retrieval times. The system will take faces as input for creating the facial embeddings for generating the databases, while the input for the recognition system is a standard video feed, where real time tagging of human faces will be done.

\section{Methodology}

\subsection{Overview}\label{AA}
In order to achieve our goals, we use Python 3.6 with OpenCV 4.1 \cite{opencv_library}. We will be using OpenCV's highly improved “deep neural networks” (dnn) module \cite{opencv_dnn} to extract faces from the images and the GoogLeNet Inception Architecture \cite{7298594} to calculate the facial embeddings. These embeddings will be saved to create the database of faces from which the face recognition system will later try to match and predict the identity of the person. In real time, for any faces which are recoginzed from the video feed, facial embeddings will be calculated and it will be matched with all the facial embeddings in the dataset. This is a searching problem where the elements don't have an order. So, we have to search linearly. This will take huge amount of time. We take this opportunity to parallelize the searching process.

\begin{figure*}[tp]
\centering
\includegraphics[width=\textwidth]{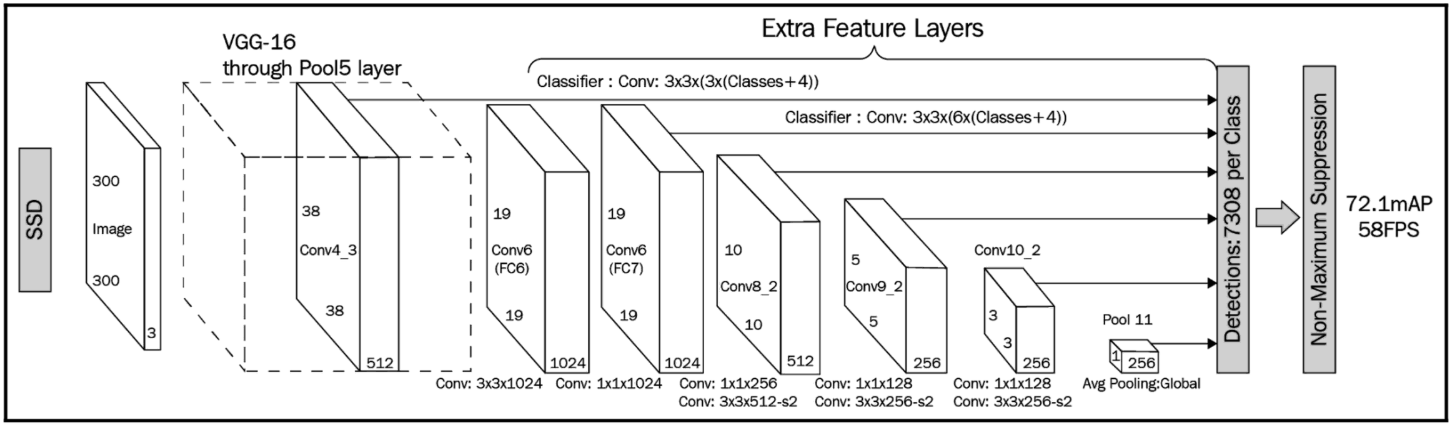}
\caption{SSD Architecture}
\label{fig:f1}
\end{figure*}

\subsection{Prerequisites}\label{AA}
\begin{enumerate}
\item A Single Shot MultiBox Detector (SSD) \cite{liu2016ssd, kapok_ssd_2018, Laganiere:2014:OCV:2692691} framework using ResNet-10 like architecture as a backbone for Face Detection. Channels count in ResNet-10 convolution layers was significantly dropped.\newline\newline
The structure of the DNN is defined as follows:
\begin{itemize}
\item A 300 x 300 input image.
\item The input image is passed through multiple convolutional layers, obtaining different features at different scales.
\item For each feature map obtained, a 3 x 3 convolutional filter to evaluate a small set of default bounding boxes.
\item For each default box evaluated, the bounding box offsets and class probabilities are predicted.
\end{itemize}
The model architecture is shown in Fig. \ref{fig:f1}

SSD’s architecture builds on the venerable VGG-16 (Visual Geometry Group) \cite{simonyan2014very} architecture, but discards the fully connected layers. The reason VGG-16 was used as the base network is because of its strong performance in high quality image classification tasks. Instead of the original VGG fully connected layers, a set of auxiliary convolutional layers (from conv6 onwards) were added, thus enabling to extract features at multiple scales and progressively decrease the size of the input to each subsequent layer.

\item Pre-trained weight values for the detection model (trained using Caffe \cite{jia2014caffe} framework).

\begin{figure*}[tp]
\centering
\includegraphics[width=\textwidth]{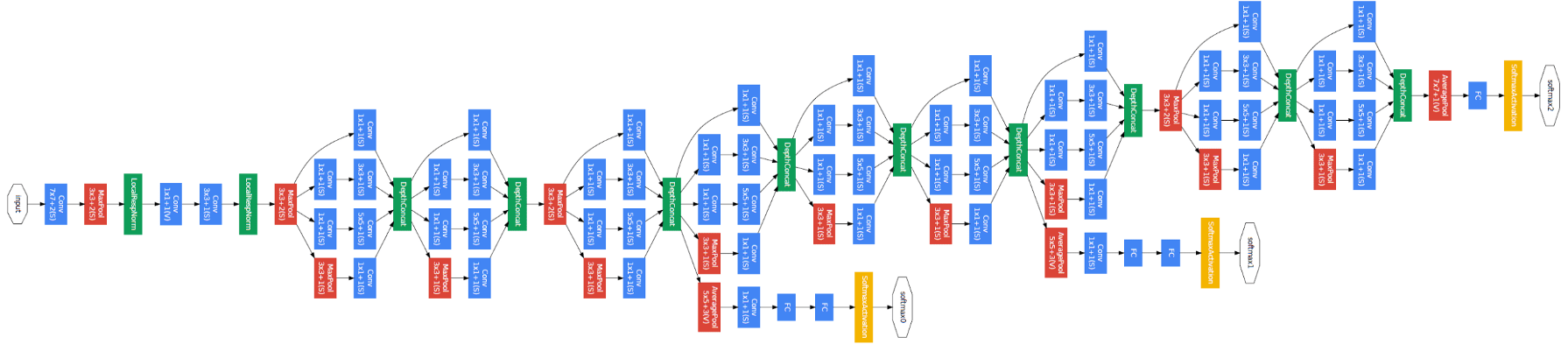}
\caption{GoogLeNet Inception Model Architecture}
\label{fig:f2}
\end{figure*}

\item GoogLeNet Inception Architecture \cite{7298594} to calculate facial embeddings. The structure of the network can be summarized as follows:
\begin{itemize}
\item An average pooling layer with 5x5 filter size and stride 3, resulting in an 4x4x512 output for the (4a), and 4x4x528 for the (4d) stage.
\item A 1x1 convolution with 128 filters for dimension reduction and rectified linear activation.
\item A fully connected layer with 1024 units and rectified linear activation.
\item A dropout layer with 70\% ratio of dropped outputs.
\item A linear layer with softmax loss as the classifier (predicting the same 1000 classes as the main classifier, but removed at inference time).
\end{itemize}
The model architecture is shown in Fig. \ref{fig:f2}

\end{enumerate}

\subsection{Locating Faces}\label{AA}
The video feed from a camera can be processed frame by frame to locate faces in each frame. For the detection of faces in a frame, we use a pre-trained ResNet model which detects faces very accurately and is faster than other frameworks \cite{990517, 10.1007/978-3-642-31686-9_46}. The architecture is written in caffe and has a separate file for pre-trained weights. If a face is present in the frame, the detector returns the (x, y) coordinates of the top-most point of the bounding box along with the height and width of the box. We managed to crop the ROI(region of interest) from the bounding box to get the face from the image.

\subsection{Facial Embeddings Calculation}\label{AA}
The cropped ROI image is the loaded and fed into the GoogLeNet Inception Model for calculation of 128-d embeddings. These embeddings represent the unique features of a person's face. If somehow, the facial embeddings of two face images are nearly the same, they are at least supposed to be similar looking if not the same person or identical twins. In order to match the facial embeddings of one face with another, euclidean distance is calculated. If the distance is smaller than a specified threshold (which should ideally be zero) we assume the faces are of the same person.

\subsection{Database Creation}\label{AA}
We can use the previously mentioned methods for locating faces and calculating facial embeddings to create a database of faces. Images of people can be used to calculate the facial embeddings which can be saved in a file or a database along with their names or ID. New entries can be appended to the file or database later or previous entries can be deleted easily.

\subsection{Real Time Face Recognition}\label{AA}
The face detection model and the facial embedding calculation used in this approach are inherently very fast but can slow down considerably when face matching has to be done on a large dataset. This is because it is a searching problem where distance of a facial embedding has to calculated with respect to all of the entries in the previously created database. Here we have an opportunity to speed up the process. Instead of comparing the distance with the threshold distance linearly (or sequentially), we can distribute the search area within the database among number of threads. When we employ multiple threads to search in different parts of the database parallely, we can find the required result faster.

\begin{algorithm}
\DontPrintSemicolon
\SetAlgoLined
\SetKwInOut{Input}{Input}
\Input{Directory containing face images and ID}
\KwResult{The Facial Embeddings Database}
\BlankLine
\BlankLine
FIND all images in the input directory\;
\ForEach{image in directory}{    
    FIND location and dimensions of faces in image\;
    \ForEach{face in image}{ 
        CALCULATE facial embeddings\;
        APPEND ID and facial embeddings in list of kmown faces\;
    }
}
SAVE contents of list of known faces\;
\caption{Algorithm for Database Creation}
\label{alg:db}
\end{algorithm}

Algorithm \ref{alg:db} creates the facial embedding database from a directory containing the images containing faces.

\begin{algorithm}
\DontPrintSemicolon
\SetAlgoLined
\SetKwInOut{Input}{Input}
\Input{Facial Embeddings Database and Live Camera Feed}
\KwResult{Identification of Daces in the Video Feed}
\BlankLine
\BlankLine
\ForEach{frame from video feed}{    
    FIND location and dimensions of faces in image\;
    \ForEach{face in image}{ 
        CALCULATE facial embeddings\;
        DISTRIBUTE list of known faces among workers\;
        \ForEach{entry in list of known faces}{ 
            CALCULATE euclidean distance from the current face\;
            APPEND distance to list of distances\;
        }
        FIND index of the smallest distance\;
        PRINT name of person at index of smallest distance\;
    }
}
\caption{Parallel Algorithm for Face Recognition}
\label{alg:fc}
\end{algorithm}

Algorithm \ref{alg:fc} takes in a video feed and uses the previously created facial embedding database to find the best match for faces in the video feed.

\begin{figure}[bhp]
\centering
\includegraphics[width=\columnwidth]{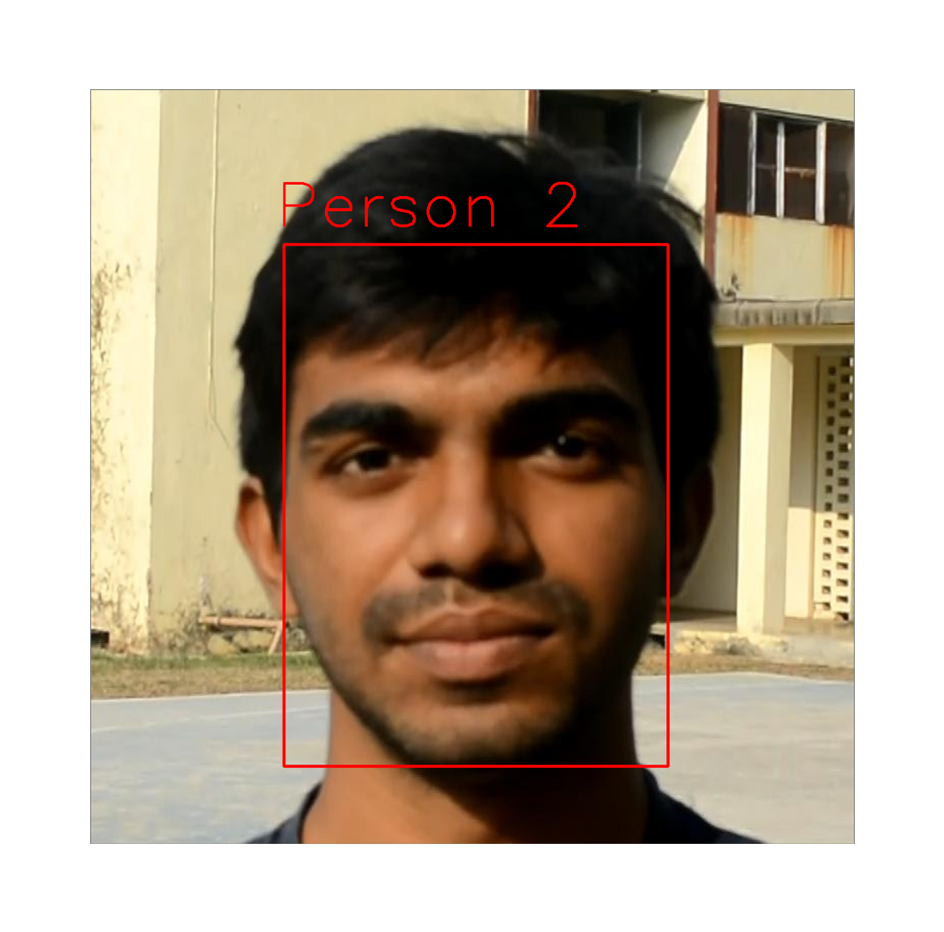}
\caption{Person identified from the QDF Dataset}
\label{fig:f3}
\end{figure}

\begin{figure}[bhp]
\centering
\includegraphics[width=\columnwidth]{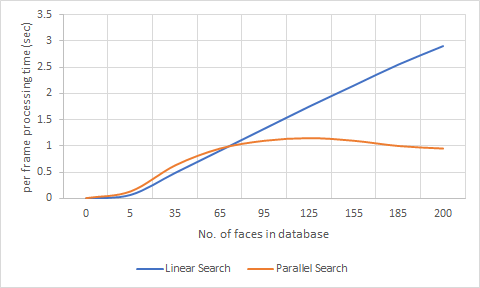}
\caption{Speedup}
\label{fig:f4}
\end{figure}

\section{Results}
The algorithm is implemented in Python 3.6 programming language and it utilizes the open source computer vision library, OpenCV. For our experimental evaluation we used a system with an Intel Core i7 6700HQ processor, with 4 cores and 8 threads. The experimentation was conducted on the QDF Dataset \cite{bhattacharya2019qdf}. The dataset contains face images of over 100 subjects in different orientations. The test was conducted with varying sizes of facial embeddings database and the results were plotted in a graph. The time to process per frame greatly reduces with larger database. 

The live output of the implementation can be seen in Fig. \ref{fig:f3}.

The speedup can be observed in Fig. \ref{fig:f4}.

\section{Conclusion}
A robust solution was created for the defined problem statement. We were able to model a solution that could store, retrieve and compare facial embeddings for the purpose of face recognition. Due to our usage of the CPU parallelization, the algorithm works more efficiently, without sacrificing any accuracy. Our model can be used for various applications, either specific or multipurpose. Primarily, it can be used for criminal identification while entry or exit at a public place. Other uses can be person wise activity monitoring system by analysing where in a building the person spends time at what time of the day. There can be many more such applications which use this approach as their base and later use it for something else.

\bibliographystyle{bibliography/IEEEtran}
\bibliography{ref.bib}

\end{document}